\documentclass[11pt,a4paper]{article}
\usepackage[hyperref]{acl2021}
\usepackage{times}
\usepackage{latexsym}
\usepackage{float}
\restylefloat{table}

\usepackage{microtype}
\usepackage{graphicx}
\usepackage{xcolor}
\usepackage{epsfig}
\usepackage{url}

\usepackage{amsmath}
\usepackage{amssymb}
\usepackage{multirow}

\aclfinalcopy 
\def\aclpaperid{1170} 

\title{\textcolor{blue}{Sa}\textcolor{yellow}{Ro}\textcolor{red}{Co}: Detecting \textcolor{blue}{Sa}tire in a Novel \textcolor{yellow}{Ro}manian \textcolor{red}{Co}rpus of News Articles}

\author{Ana-Cristina Rogoz, Mihaela G\u{a}man, Radu Tudor Ionescu* \\
  University of Bucharest\\
  14 Academiei Street, Bucharest, Romania \\
  \texttt{*raducu.ionescu@gmail.com}}

\date{}

\begin{document}
\maketitle

\begin{abstract}
In this work, we introduce a corpus for satire detection in Romanian news. We gathered 55,608 public news articles from multiple real and satirical news sources, composing one of the largest corpora for satire detection regardless of language and the only one for the Romanian language. We provide an official split of the text samples, such that training news articles belong to different sources than test news articles, thus ensuring that models do not achieve high performance simply due to overfitting. We conduct experiments with two state-of-the-art deep neural models, 
resulting in a set of strong baselines for our novel corpus. Our results show that the machine-level accuracy for satire detection in Romanian is quite low (under $73\%$ on the test set) compared to the human-level accuracy ($87\%$), leaving enough room for improvement in future research.
\end{abstract}

\section{Introduction}
\label{sec_intro}

According to its definition in the Cambridge Dictionary, satire is ``a humorous way of criticizing people or ideas''
. News satire employs this mechanism in the form of seemingly legitimate journalistic reporting, with the intention of ridiculing public figures, politics or contemporary events \cite{McClennen-Springer-2014, Peters-Routledge-2013,Rubin-CADD-2016}. Although the articles pertaining to this genre contain fictionalized stories, the intent is not to mislead the public into thinking that the discussed subjects are real. On the contrary, satirical news articles are supposed to reveal their nature by the writing style and comedic devices employed, such as irony, parody or exaggeration. Thus, the intention behind the writing differentiates satirical news \cite{Rubin-CADD-2016} from fake news \cite{Meel-ESA-2019,Perez-COLING-2018,Sharma-TIST-2019}. However, in some rare cases, the real intent might be deeply buried in the complex irony and subtleties of news satire \cite{Barbieri-IJCAI-2015}, which has the effect of fiction being deemed as factual \cite{Zhang-ArXiv-2020}. Even so, there is a clear distinction between satirical and fake news. In fake news, the intent is to deceive the readers in thinking that the news is real, while presenting fake facts to influence the readers' opinion. Since our study is focused on satire detection, we consider discussing research on fake news detection as being out of our scope. 
At the same time, we acknowledge the growing importance of detecting fake news and the fact that an accurate differentiation of satirical from legitimate journalistic reports might be seen as a starting point in controlling the spread of deceptive news \cite{DeSarkar-COLING-2018}.

\begin{table*}[t]
\begin{center}
\footnotesize
\begin{tabular}{|l|l|r|r|r|}
\hline
\multirow{2}{*}{Data Set}					& \multirow{2}{*}{Language}      & \multicolumn{3}{|c|}{\#articles}  \\
\cline{3-5}
     						&               & Regular       & Satirical	        &  Total\\
\hline
\hline
\cite{Burfoot-ACL-2009}     & English       & 4,000			& 233		        & 4,233 \\
\cite{Frain-LREC-2016}      & English       & 1,705			& 1,706		        & 3,411 \\
\cite{Goldwasser-TACL-2016} & English       & 10,921		& 1,225		        & 12,146 \\
\cite{Ionescu-Arxiv-2021}   & French        & 5,648         & 5,922             & 11,570 \\
\cite{Li-NLP4IF-2020}       & English       & 6,000			& 4,000		        & 10,000 \\
\cite{Liu-ICCCN-2019}       & French        & 2,841         & 2,841             & 5,682 \\
\cite{McHardy-NAACL-2019}   & German        & 320,219 		& 9,643 		    & 329,862 \\
\cite{Ravi-KBS-2017}		& English       & 1,272			& 393		        & 1,665 \\
\cite{Saadany-RDSM-2020}    & Arabic        & 3,185			& 3,710		        & 6,895 \\
\cite{Tocoglu-ID-2019}      & Turkish       & 1,000			& 1,000		        & 2,000 \\
\cite{Yang-EMNLP-2017}      & English       & 168,780		& 16,249		    & 185,029 \\
\hline
SaRoCo (ours)				& Romanian      & 27,980		& 27,628		    & 55,608  \\
\hline
\end{tabular}
\end{center}
\caption{Number of regular and satirical news articles in existing corpora versus SaRoCo.}
\label{tab_corpora}
\end{table*}


\begin{table*}[t]
\begin{center}
\footnotesize
\begin{tabular}{|l|r|r|r|r|r|r|}
\hline
\multirow{2}{*}{Set} 						& \multicolumn{2}{|c|}{Regular}     & \multicolumn{2}{|c|}{Satirical}   & \multicolumn{2}{|c|}{Total}\\
\cline{2-7}
     						& \#articles		& \#tokens	        & \#articles		& \#tokens          & \#articles		& \#tokens\\
\hline
\hline
Training				    & 18,000 		& 8,174,820         & 17,949 		& 11,147,169         & 35,949        & 19,321,989\\
Validation					& 4,986			& 2,707,026           & 4,878		    & 3,030,055           & 9,864         & 5,737,081\\
Test					    & 4,994			& 2,124,346           & 4,801		    & 1,468,199           & 9,795         & 3,592,545\\
\hline
Total						& 27,980		& 13,006,192         & 27,628		& 15,645,423         & 55,608        & 28,651,615\\
\hline
\end{tabular}
\end{center}
\caption{Number of samples (\#articles) and number of tokens (\#tokens) for each subset in SaRoCo.}
\label{tab_SaRoCo}
\end{table*}

Satire detection is an important task that could be addressed prior to the development of conversational systems and robots that interact with humans. Certainly, the importance of understanding satirical (funny, ridiculous or ironical) text becomes obvious when we consider a scenario in which a robot performs a dangerous action because it takes a satirical comment of the user too literally. Given the relevance of the task for the natural language processing community, satire detection has already been investigated in several well-studied languages such as Arabic \cite{Saadany-RDSM-2020}, English \cite{Burfoot-ACL-2009,DeSarkar-COLING-2018,Goldwasser-TACL-2016,Yang-EMNLP-2017}, French \cite{Ionescu-Arxiv-2021,Liu-ICCCN-2019}, German \cite{McHardy-NAACL-2019}, Spanish \cite{Barbieri-PLN-2015} and Turkish \cite{Tocoglu-ID-2019}. Through the definition of satire, the satire detection task is tightly connected to irony and sarcasm detection. These tasks strengthen or broaden the language variety with languages such as Arabic \cite{Karoui2017}, Chinese \cite{Jia2019}, Dutch \cite{Liebrecht2013} and Italian \cite{Giudice-EVALITA-2018}. 

In this work, we introduce SaRoCo\footnote{\url{https://github.com/MihaelaGaman/SaRoCo}}, the \textbf{Sa}tire detection \textbf{Ro}manian \textbf{Co}rpus, which comprises 55,608 news articles collected from various sources. To the best of our knowledge, this is the first and only data set for the study of Romanian satirical news. Furthermore, SaRoCo is also one of the largest data sets for satirical news detection, being surpassed only two corpora, one for English \cite{Yang-EMNLP-2017} and one for German \cite{McHardy-NAACL-2019}. However, our corpus contains the largest collection of satirical news articles (over 27,000). These facts are confirmed by the comparative statistics presented in Table~\ref{tab_corpora}.

Along with the novel data set, we include two strong deep learning methods to be used as baselines in future works. The first method is based on low-level features learned by a character-level convolutional neural network \cite{Zhang-NIPS-2015}, while the second method employs high-level semantic features learned by the Romanian version of BERT \cite{Dumitrescu-EMNLP-2020}. The gap between the human-level performance and that of the deep learning baselines 
indicates that there is enough room for improvement left for future studies. We make our corpus and baselines available online for nonprofit educational and research purposes, under an open-source noncommercial license agreement.



\section{Corpus}
\label{sec:benchmark}

SaRoCo gathers both satirical and non-satirical news from some of the most popular Romanian news websites. The collected news samples were found in the public web domain, i.e.~access is provided for free without requiring any subscription to the publication sources. The entire corpus consists of 55,608 samples (27,628 satirical samples and 27,980 non-satirical samples), having more than 28 million tokens in total, as illustrated in Table \ref{tab_SaRoCo}. Each sample is composed of a title (headline), a body and a corresponding label (satirical or non-satirical). As shown in Table~\ref{tab_avg_stats}, an article has around $515.24$ tokens on average, with an average of $24.97$ tokens for the headline. We underline that the labels are automatically determined, based on the fact that a publication source publishes either regular or satirical news, but not both.

\begin{table}[!t]
\footnotesize
\begin{center}
 \begin{tabular}{|l | r |} 
 \hline
 Sample Part    & Average \#tokens\\
 \hline 
 \hline
 Title          & 24.97 \\ 
 Full Articles  & 515.24 \\
 \hline
\end{tabular}
\end{center}
\caption{Average number of tokens in full news articles and titles from SaRoCo.}
\label{tab_avg_stats}
\end{table}

We provide an official split for our corpus, such that all future studies will use the same training, validation and test sets, easing the direct comparison with prior results. Following \newcite{McHardy-NAACL-2019}, we use disjoint sources for training, validation and test, ensuring that models do not achieve high performance by learning author styles or topic biases particular to certain news websites. While crawling the public news articles, we selected the same topics (culture, economy, politics, social, sports, tech) and the same time frame (between 2011 and 2020) for all news sources to control for potential biases induced by uneven topic or time distributions across the satirical and non-satirical genres.

\begin{table}[t]
\setlength\tabcolsep{4.8pt}
\centering
\footnotesize
\begin{tabular}{|p{0.16\linewidth}|p{0.34\linewidth}|p{0.35\linewidth}|}
\hline
Category &  Example & Translation \\
\hline 
\hline
\multirow{13}{*}{Regular}            & \textit{``Tragedie \^{i}n zi de s\u{a}rb\u{a}toare''} & \textit{``Tragedy during celebration day''} \\
\cline{2-3}
            & \textit{``Demisia lui \$NE\$ \$NE\$ se am\^{a}n\u{a}''} & \textit{``\$NE\$ \$NE\$'s resignation is post-poned''} \\
\cline{2-3}
            & \textit{``Premierul bulgar \$NE\$ \$NE\$ are \$NE\$''} & \textit{``Bulgarian prime-minister \$NE\$ \$NE\$ has \$NE\$''} \\
\cline{2-3}
            & \textit{``A murit actorul \$NE\$ \$NE\$''} & \textit{``The actor \$NE\$ \$NE\$ died''} \\
\cline{2-3}
            & \textit{``Metroul din \$NE\$ \$NE\$ se deschide azi''} & \textit{``Subway to \$NE\$ \$NE\$ opens up today''} \\
\hline
\multirow{15}{*}{Satirical}        & \textit{``Comedia cu p\u{a}l\u{a}rioar\u{a} de staniol''} &  \textit{``Comedy with little tinfoil hat''} \\
\cline{2-3}
        & \textit{``10 restric\c{t}ii dure pe care \$NE\$ le preg\u{a}te\c{s}te pe ascuns''} & \textit{``10 harsh restrictions that \$NE\$ is planning in secrecy''} \\
\cline{2-3}
 & \textit{``C\^{a}\c{t}i pokemoni ai prins azi?''} & \textit{``How many pokemons did you catch today?''}\\
\cline{2-3}        
        & \textit{``Biserica \$NE\$ lanseaz\u{a} apa sfin\c{t}it\u{a} cu arom\u{a}''} & \textit{``The \$NE\$ Church launches flavored holy water''}\\
\cline{2-3}
        & \textit{``Dragostea \^{i}n vremea sclerozei''} & \textit{``Love in the time of sclerosis''} \\
\hline
\end{tabular}
\caption{Examples of news headlines from SaRoCo.}
\label{tab_examples_news_headlines}
\end{table}

\begin{table*}[!t]
\setlength\tabcolsep{4.2pt}
\begin{center}
\footnotesize
\begin{tabular}{|l|c|c|c|c|c|c|c|c|c|c|c|c|c|c|c|c|}
\hline 
\multirow{3}{*}{Method} & \multicolumn{6}{|c|}{Validation} & \multicolumn{6}{|c|}{Test} \\
\cline{2-13}
 						&  \multirow{2}{*}{Acc.} & Macro & \multicolumn{2}{|c|}{Satirical} & \multicolumn{2}{|c|}{Regular} & \multirow{2}{*}{Acc.} & Macro & \multicolumn{2}{|c|}{Satirical} & \multicolumn{2}{|c|}{Regular} \\
\cline{4-7} \cline{10-13}
&  & $F_1$ & Prec. & Rec. & Prec. & Rec. & & $F_1$ & Prec. & Rec. & Prec. & Rec.\\
\hline
\hline
Ro-BERT  & 0.8241	& 0.8160	& 0.9260	& 0.6991	& 0.7633	& 0.9462 &	0.7300	& 0.7150 &	0.8750 &	0.5250 &	0.6700 &	0.9250 \\
Char-CNN & 0.7342 & 0.7475 & 0.8023 & 0.6138 & 0.6928 & 0.8520 & 0.6966 & 0.7109 & 0.7612 & 0.5551 & 0.6606 & 0.8326 \\
\hline
\end{tabular}
\end{center}
\caption{Validation and test results of the character-level CNN and the fine-tuned Ro-BERT applied on SaRoCo.}
\label{tab_results_full}
\end{table*}

After crawling satirical and non-satirical news samples, our first aim was to prevent discrimination based on named entities. The satirical character of an article should be inferred from the language use rather than specific clues, such as named entities. For example, certain sources of news satire show preference towards mocking politicians from a specific political party, and an automated system might erroneously label a news article about a member of the respective party as satirical simply based on the presence of the named entity. Furthermore, we even noticed that some Romanian politicians have certain mocking nicknames assigned in satirical news. In order to eliminate named entities, we followed a similar approach as the one used for the MOROCO \cite{Butnaru-ACL-2019} data set. Thus, all the identified named entities are replaced with the special token \$NE\$.
Besides eliminating named entities, we also substituted all whitespace characters with space and replaced multiple consecutive spaces with a single space. A set of processed satirical and regular headlines are shown in Table~\ref{tab_examples_news_headlines}.






\section{Baselines}

\noindent
\textbf{Fine-tuned Ro-BERT.}
Our first baseline consists of a fine-tuned Romanian BERT \cite{Dumitrescu-EMNLP-2020}, which follows the same transformer-based model architecture as the original BERT \cite{Devlin-NAACL-2019}. According to \newcite{Dumitrescu-EMNLP-2020}, the Romanian BERT (Ro-BERT) attains better results than the multilingual BERT on a range of tasks. We therefore assume that the Romanian BERT should represent a stronger baseline for our Romanian corpus.

We use the Ro-BERT encoder to encode each text sequence into a list of token IDs. The tokens are further processed by the model, obtaining the corresponding 768-dimensional embeddings. At this point, we add a global average pooling layer to obtain a Continuous Bag-of-Words (CBOW) representation for each sequence of text, followed by a Softmax output layer with two neural units, each predicting the probability for one category, either non-satirical or satirical. To obtain the final class label for a text sample, we apply \emph{argmax} on the two probabilities. We fine-tune the whole model for 10 epochs on mini-batches of 32 samples, using the Adam with decoupled weight decay (AdamW) optimizer \cite{Loshchilov-ICLR-2019}, with a learning rate of $10^{-7}$ and the default value for $\epsilon$.

\noindent
\textbf{Character-level CNN.}
The second baseline model considered in the experiments is a Convolutional Neural Network (CNN) that operates at the character level \cite{Zhang-NIPS-2015}. We set the input size to 1,000 characters. 
After the input layer, we add an embedding layer to encode each character into a vector of $128$ components. The optimal architecture for the task at hand proved to be composed of three convolutional (conv) blocks, each having a conv layer with $64$ filters applied at stride $1$, followed by Scaled Exponential Linear Unit (SELU) activation. From the first block to the third block, the convolutional kernel sizes are $5$, $3$ and $1$, respectively. Max-pooling with a filter size of $3$ is applied after each conv layer. After each conv block, we insert a Squeeze-and-Excitation block with the reduction ratio set to $r = 64$, following \newcite{Butnaru-ACL-2019}. To prevent overfitting, we use batch normalization and Alpha Dropout \cite{Klambauer-NIPS-2017} with a dropout rate of 
$0.5$. The final prediction layer is composed of two neural units, one for each class (i.e.~legitimate and satirical), with Softmax activation. We use the Nesterov-accelerated Adaptive Moment Estimation (Nadam) optimizer \cite{Dozat-ICLR-2016} with a learning rate of $2\cdot10^{-4}$, training the network for $50$ epochs on mini-batches of $128$ samples.

\section{Experiments}

\noindent
\textbf{Evaluation.}
We conducted binary classification experiments on SaRoCo, predicting if a given piece of text is either satirical or non-satirical. As evaluation metrics, we employ the precision and recall for each of the two classes. We also combine these scores through the macro $F_1$ and micro $F_1$ (accuracy) measures.

\noindent
\textbf{Results.}
In Table~\ref{tab_results_full}, we present the results of the two baselines on the SaRoCo validation and test sets. We observe that both models tend to have higher precision scores in detecting satire than in detecting regular news. The trade-off between precision and recall is skewed towards higher recall for the non-satirical news class. Since both models share the same behavior, we conjecture that the behavior is rather caused by the particularities of the satire detection task.

\begin{table}[!t]
\setlength\tabcolsep{4.8pt}
\centering
\footnotesize
\begin{tabular}{|p{0.2\linewidth}|p{0.31\linewidth}|p{0.33\linewidth}|}
\hline
{Category} &  Example & Translation \\
\hline 
\hline
Slang & \textit{``cel mai marf\u{a} serial din lume''} & \textit{``the dopest TV show in the world''} \\
                     & \textit{``cocalar''} & \textit{``douche''} \\
\hline
Insult & \textit{``odiosul primar''} & \textit{``the odious mayor''}\\
                      & \textit{``bunicu\c{t} retardat''} & \textit{``retarded grandpa''}\\
                      & \textit{``dugongul \u{a}la slinos de la sectorul 4''} & \textit{``that slender dugong in the 4th sector''}\\
\hline
Repetition & \textit{``Mii de gunoaie care las\u{a} gunoaie au remarcat c\u{a} [...] plajele [...] s-au umplut de gunoaie, l\u{a}sate [...] de gunoaiele care au venit \^{i}naintea lor''} & \textit{``Thousands of scums who leave garbage noticed that [...] beaches [...] got full of garbage, left behind [...] by the scums who were there before them''}\\
\hline
Exaggeration    & \textit{``Ne-am s\u{a}turat!''}                   & \textit{``We're sick of it!''}\\
Exclamation    & \textit{``Ru\c{s}ine s\u{a} le fie!''}                & \textit{``Shame on them!''}\\
Irony           & \textit{``Chiar nu suntem o na\c{t}ie de ho\c{t}i!''} & \textit{``We're totally not a nation of thieves!''}\\
\hline
Popular & \textit{``a s\u{a}rit calul''}     & \textit{``went overboard''}\\
Saying & \textit{``a f\u{a}cut-o de oaie''} & \textit{``messed up''}\\
                      & \textit{``minte de g\u{a}in\u{a}''}    & \textit{``bird brain''}\\
\hline
\end{tabular}
\caption{Examples of predictive patterns of satire learned by the character-level CNN.}
\label{tab_most_predictive_patterns_satire}
\end{table}


\begin{table}[!t]
\setlength\tabcolsep{4.8pt}
\centering
\footnotesize
\begin{tabular}{|p{0.2\linewidth}|p{0.31\linewidth}|p{0.33\linewidth}|}
\hline
Category &  Example & Translation \\
\hline 
\hline
Stats & \textit{``Importurile au sc\u{a}zut cu $2.1\%$ [...] pentru o cre\c{s}tere de $0.1\%$ \c{s}i prelungirea sc\u{a}derii de $1.4\%$ din iulie.''} & \textit{``Imports decreased by $2.1\%$ [...] for an increase of $0.1\%$ and the prolongation of the decrease of $1.4\%$ since July.''} \\
\hline
Legal terms   & \textit{``asasinat''} & \textit{``assasinated''} \\
                        & \textit{``l-au denun\c{t}at pe autorul atacului''} & \textit{``denounced the perpetrator''} \\
\hline
Weather    & \textit{``temperatura \^{i}n timpul nop\c{t}ii a sc\u{a}zut''} & \textit{``the temperature has dropped during the night''} \\
\hline
Political terms     & \textit{``scrutinul preziden\c{t}ial''} & \textit{``presidential election''} \\
                    & \textit{``prefectura informeaz\u{a} c\u{a}''} & \textit{``prefecture informs that''} \\
\hline
\end{tabular}
\caption{Examples of predictive patterns of legitimate news learned by the character-level CNN.}
\label{tab_most_predictive_patterns_legitimate}

\end{table}

\noindent
\textbf{Discriminative feature analysis.}
We analyze the discriminative features learned by the character-level CNN, which is one of the proposed baseline systems for satire detection. We opted for the character-level CNN in favor of the fine-tuned BERT, as the former method allows us to visualize discriminative features using Grad-CAM~\cite{Selvaraju-ICCV-2017}, a technique that was initially used to explain decisions of CNNs applied on images. We adapted this technique for the character-level CNN, then extracted and analyzed the most predictive patterns in SaRoCo. The motivation behind this was to validate that the network's decisions are not based on some biases that escaped our data collection and cleaning process. 

\begin{table*}[!t]
\centering
\footnotesize
\begin{tabular}{|l|c|c|c|c|c|c|}
\hline
\multirow{2}{*}{Method} & \multirow{2}{*}{Acc.} & Macro & \multicolumn{2}{|c|}{Satirical} & \multicolumn{2}{|c|}{Regular} \\
\cline{4-7}
& & $F_1$ & Prec. & Rec. & Prec. & Rec.\\
\hline
\hline 
Ro-BERT & 0.6800	& 0.6750 &	0.7800 &	0.5100 &	0.6350 &	0.8550 \\
Char-CNN & 0.6500 &  0.6510 & 0.6389 & 0.6900 & 0.6630 & 0.6100 \\
\hline
Humans & 0.8735 & 0.8711 & 0.9416 & 0.7970 & 0.8332 & 0.9500 \\
\hline
\end{tabular}
\caption{Averaged performance of ten human annotators versus deep learning baselines on 200 news headlines from SaRoCo. 
}
\label{tab_human_eval}
\end{table*}

In Tables~\ref{tab_most_predictive_patterns_satire} and~\ref{tab_most_predictive_patterns_legitimate}, we present a few examples of interesting patterns considered relevant for predicting satire versus regular news, respectively. A broad range of constructions covering a great variety of styles and significant words are underlined via Grad-CAM in the satirical news samples. The network seems to pick up obvious clues such as slang, insults and popular sayings rather than more subtle indicatives of satire, including irony or exaggeration. 
At the same time, for the real news in SaRoCo, there are fewer categories of predictive patterns. In general, the CNN deems formal, standard news expressions as relevant for regular news. These patterns vary across topics and domains. The CNN also finds that the presence of numbers and statistical clues is indicative for non-satirical content, which is consistent with the observations of \newcite{Yang-EMNLP-2017}. Our analysis reveals that the discriminative features are appropriate for satire detection, showing that our corpus is indeed suitable for the considered task.

\noindent
\textbf{Deep models versus humans.}
Given 100 satirical and 100 non-satirical news headlines (titles) randomly sampled from the SaRoCo test set, we asked ten Romanian human annotators to label each sample as satirical or non-satirical. We evaluated the deep learning methods on the same subset of 200 samples, reporting the results in Table~\ref{tab_human_eval}. First, we observe that humans have a similar bias as the deep learning models. Indeed, for both humans and models, the trade-off between precision and recall is skewed towards higher precision for the satirical class and higher recall for the non-satirical class. We believe this is linked to the way people and machines make a decision. Humans look for patterns of satire in order to label a sample as satire. If a satire-specific pattern is not identified, the respective sample is labeled as regular, increasing the recall for the non-satirical class. Although humans and machine seem to share the same way of thinking, there is a considerable performance gap in satire detection between humans and machines. Indeed, the average accuracy of our ten human annotators is around $87\%$, while the state-of-the-art deep learning models do not surpass $68\%$ on the same news headlines. Even on full news articles (see Table~\ref{tab_results_full}), the models barely reach an accuracy of $73\%$ on the test set. Hence, we conclude there is a significant performance gap between humans and machines, leaving enough room for exploration in future work on Romanian satire detection.

We would like to emphasize that our human evaluation was performed by casual news readers, and the samples were shown after named entity removal, thus having a fair comparison with the AI models. We underline that named entity removal makes the task more challenging, even for humans.

\section{Conclusion}

In this work, we presented SaRoCo, a novel data set containing satirical and non-satirical news samples. To the best of our knowledge, SaRoCo is the only corpus for Romanian satire detection and one of the largest corpora regardless of language. We trained two state-of-the-art neural models as baselines for future research on our novel corpus. We also compared the performance of the neural models with the averaged performance of ten human annotators, showing that the neural models lag far behind the human-level performance. Our discriminative feature analysis confirms the limitations of state-of-the-art neural models in detecting satire. Although we selected a set of strong models from the recent literature as baselines for SaRoCo, significant future research is necessary to close the gap with respect to the human-level satire detection performance. Designing models to pick up irony or exaggerations could pave the way towards closing this gap in future work.

\section*{Acknowledgments}

The authors thank reviewers for their useful remarks. 
This work was supported by a grant of the Ministry of Research, Innovation and Digitization, CNCS/CCCDI – UEFISCDI, project number PN-III-P1-1.1-TE-2019-0235, within PNCDI III. This article has also benefited from the support of the Romanian Young Academy, which is funded by Stiftung Mercator and the Alexander von Humboldt Foundation for the period 2020-2022.



\bibliographystyle{acl_natbib}
\bibliography{acl2021-arxiv}


\end{document}